\documentclass[letterpaper]{article} 
\usepackage{aaai2026}  
\usepackage{times}  
\usepackage{helvet}  
\usepackage{courier}  
\usepackage[hyphens]{url}  
\usepackage{graphicx} 
\usepackage{natbib}  
\usepackage{caption} 
\usepackage{algorithm}
\usepackage{algorithmic}

\usepackage{booktabs}
\usepackage{amsfonts}
\usepackage{multicol}
\usepackage{multirow}
\usepackage{makecell}
\usepackage{arydshln}
\usepackage{array}
\usepackage{bm}
\usepackage{amsmath}
\usepackage{subfig}
\usepackage{dblfloatfix}

\usepackage{pifont}  
\newcommand{\cmark}{\ding{52}}  
\newcommand{\xmark}{\ding{53}}  
\newcommand{\correct}{\textcolor{green!60!black}{\cmark}}
\newcommand{\incorrect}{\textcolor{red!80!black}{\xmark}}

\newcommand{\AXE}{AXE }
\newcommand{\axe}{AXE}
\newcommand{\hidek}[1]{}

\usepackage{newfloat}
\usepackage{listings}
\DeclareCaptionStyle{ruled}{labelfont=normalfont,labelsep=colon,strut=off} 
\floatstyle{ruled}
\newfloat{listing}{tb}{lst}{}
\floatname{listing}{Listing}
\title{Evaluating the Ability of Explanations to Disambiguate Models in a Rashomon Set}
\author{
    Kaivalya Rawal\textsuperscript{\rm 1},
    Eoin Delaney\textsuperscript{\rm 2},
    Zihao Fu\textsuperscript{\rm 3},
    Sandra Wachter\textsuperscript{\rm 1,4},
    Chris Russell\textsuperscript{\rm 1}
}
\affiliations{
    \textsuperscript{\rm 1}Oxford Internet Institute, University of Oxford\\
    \textsuperscript{\rm 2}Trinity College Dublin\\
    \textsuperscript{\rm 3}Chinese University of Hong Kong\\
    \textsuperscript{\rm 4}Hasso Plattner Institute\\
    kaivalya.rawal@oii.ox.ac.uk, eoin.delaney@tcd.ie, zihaofu@cuhk.edu.hk, sandra.wachter@hpi.de, chris.russell@oii.ox.ac.uk
}

\usepackage{bibentry}

\begin{document}

\maketitle

\begin{abstract}
Explainable artificial intelligence (XAI) is concerned with producing explanations indicating the inner workings of models. For a Rashomon set of similarly performing models, explanations provide a way of disambiguating the behavior of individual models, helping select models for deployment. However explanations themselves can vary depending on the explainer used, and need to be evaluated. In the paper ``Evaluating Model Explanations without Ground Truth'', we proposed three principles of explanation evaluation and a new method ``\axe'' to evaluate the quality of feature-importance explanations. We go on to illustrate how evaluation metrics that rely on comparing model explanations against ideal ground truth explanations obscure behavioral differences within a Rashomon set. Explanation evaluation aligned with our proposed principles would highlight these differences instead, helping select models from the Rashomon set. The selection of alternate models from the Rashomon set can maintain identical predictions but mislead explainers into generating false explanations, and mislead evaluation methods into considering the false explanations to be of high quality. \axe, our proposed explanation evaluation method, can detect this adversarial fairwashing of explanations with a 100\% success rate. Unlike prior explanation evaluation strategies such as those based on model sensitivity or ground truth comparison, \AXE can determine when protected attributes are used to make predictions. 
\end{abstract}

\begin{links}
    \link{Previously Reviewed Extended Version}{https://dl.acm.org/doi/10.1145/3715275.3732219}
    \link{Code and Datasets}{https://github.com/KaiRawal/Evaluating-Model-Explanations-without-Ground-Truth}
\end{links}

\section{Introduction}

\label{sec:introduction}

Models that make similar predictions on a given domain of input datapoints form a Rashomon set of models. These models can differ in their internal decision-making mechanisms, causing differences in critical aspects such as fairness towards protected groups or the use of protected input features. A set of models forming a Rashomon set might have similar predictions in a certain input manifold, but have divergent behavior in a different region of input datapoints. With the increasingly complex and opaque inner workings of artificial intelligence systems, the internal mechanisms underlying models often need to be discovered through dedicated explainable artificial intelligence (XAI) techniques. 
Explainability techniques can help disambiguate models from a Rashomon set with otherwise indistinguishable outputs, and aid in model selection by operationalizing criteria beyond model accuracy.

While there are many competing explanation types, data modalities, and evaluation desiderata, \cite{DBLP:journals/corr/abs-2110-10790}, this paper focuses exclusively on local feature-importance explanations for models operating on tabular datasets, such as LIME \cite{lime} or SHAP \cite{shap}. These can operate on any model type and explain individual predictions rather than describing global model behavior, producing
feature importances as output: a signed vector indicating the relative contribution of each input feature to the model prediction. 

\begin{figure}[h!]
    \centering
    \includegraphics[width=0.98\linewidth]{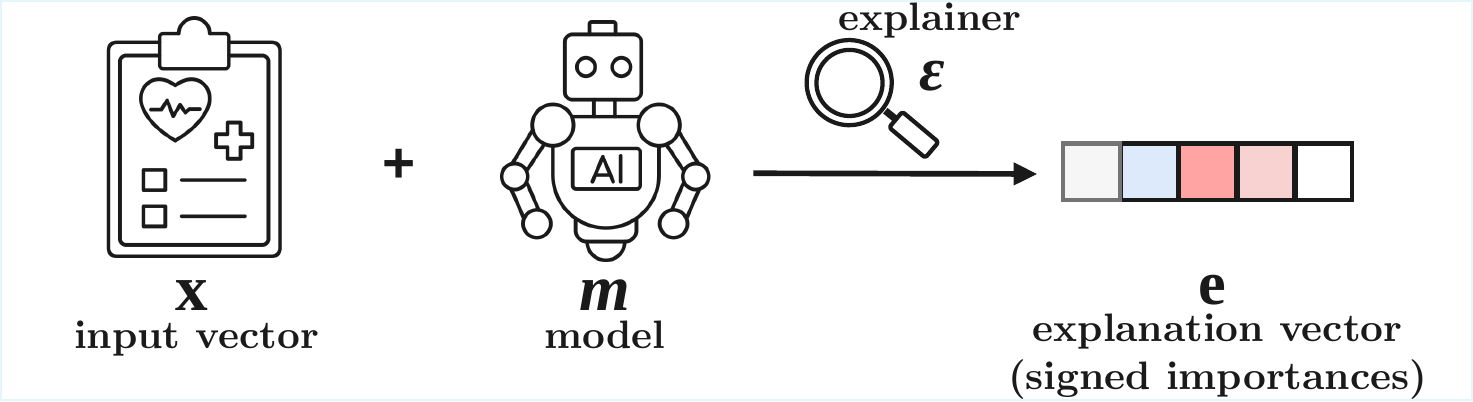}
    \caption{\textbf{Explanation generation}: Explainer $\mathcal{E}$ produces an explanation vector $\mathbf{e}$ of signed feature importances using datapoint $\mathbf{x}$, model $m$ and prediction $m(\mathbf{x})$.}
    \label{fig:infographic1}
\end{figure}

\begin{figure*}[ht]
    \centering
    \subfloat[\textbf{``Gradients''} explainer: \textbf{Diabetes Pedigree Function} is the most important input, pushing the model to a positive prediction (diabetic)]{
        \includegraphics[width=0.48\linewidth]{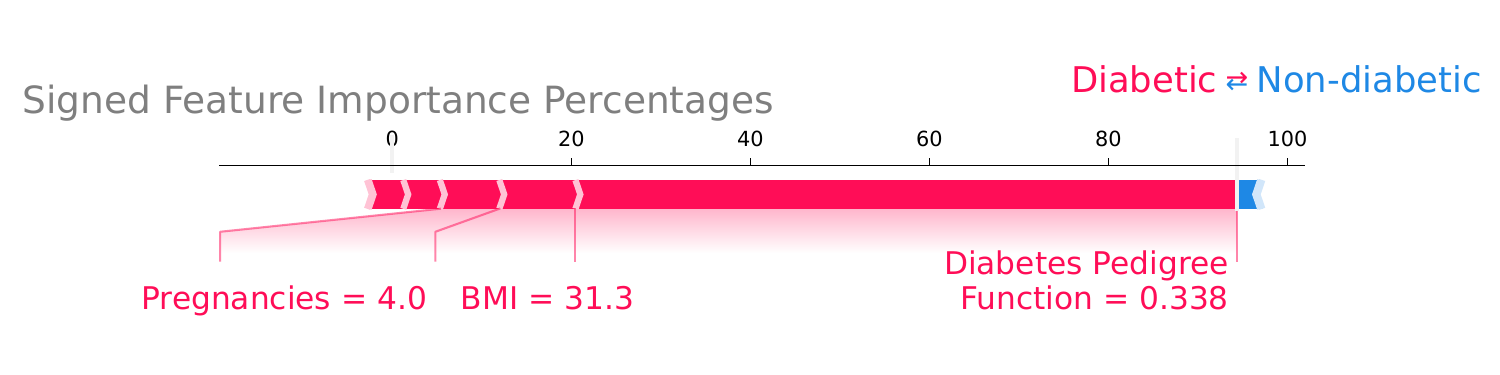}
    }
    \hfill
    \subfloat[\textbf{``SHAP''} explainer: \textbf{Glucose} is still the most important (positive) input, \textbf{BMI} now has an negative importance (non-diabetic)]{
        \includegraphics[width=0.48\linewidth]{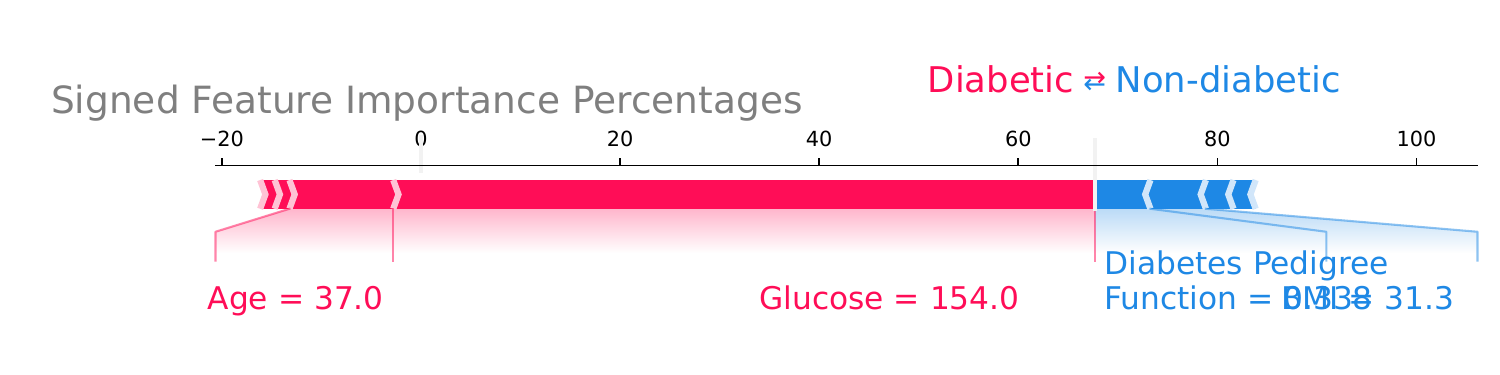}
    } \\
    \hfill
    \subfloat[\textbf{``LIME''} explainer: \textbf{Glucose} (positive), \textbf{BMI} (positive), and \textbf{Insulin} (negative) are all important input features]{
        \includegraphics[width=0.48\linewidth]{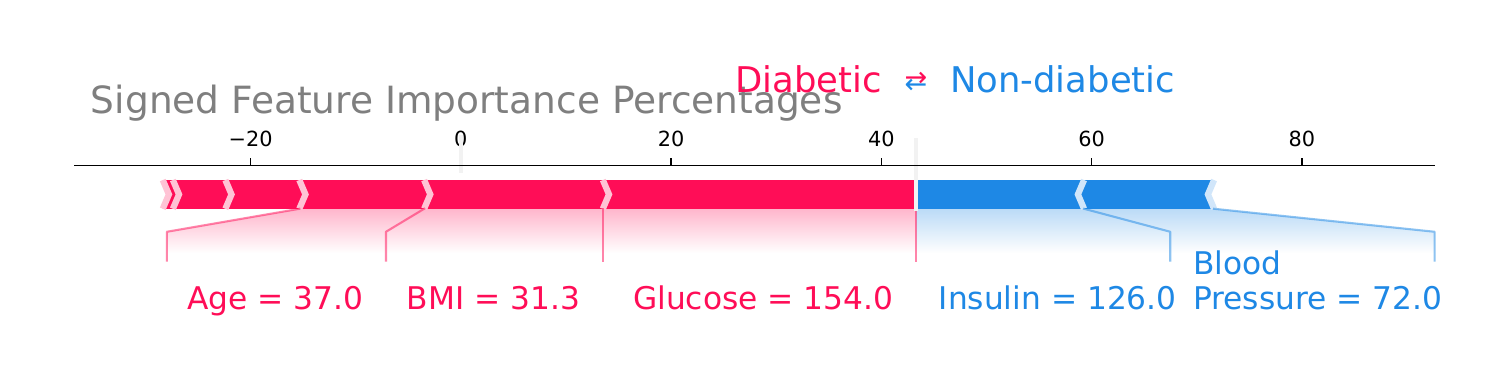}
    }
    \hfill
    \subfloat[\textbf{``Integrated Gradients''} explainer: \textbf{Glucose} (positive) \textbf{BMI} (positive) and \textbf{Blood Pressure} (negative) are important.]{
        \includegraphics[width=0.48\linewidth]{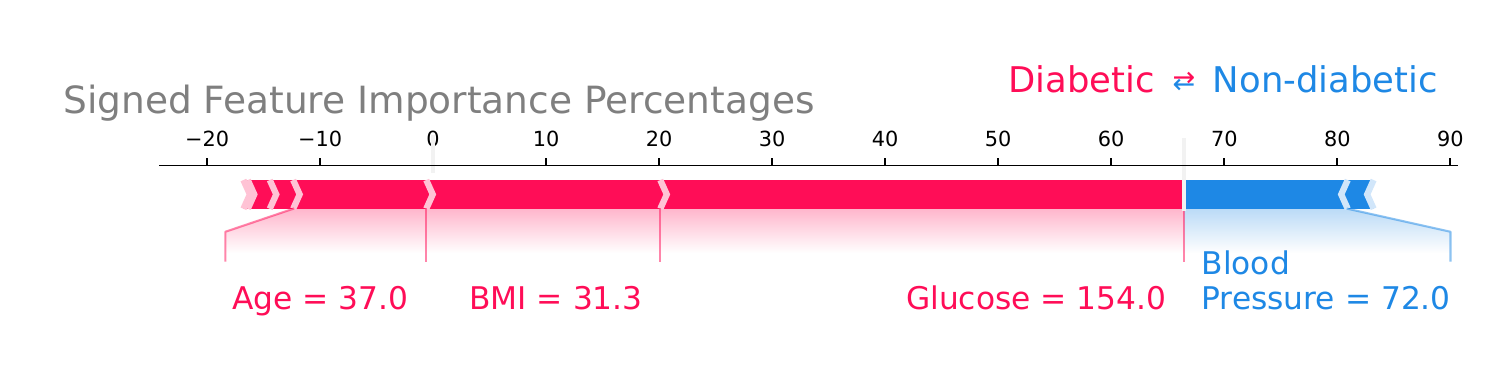}
    }
     \caption{\textbf{Different Explainers Yield Different Explanations}: A neural network predicts diabetes on the ``Pima Indians'' dataset \cite{pima}. A single positive (diabetic) prediction is explained using four explainers. Each feature-importance explanation varies, but consists of a signed vector indicating the relative contribution of each input to the model output.}
    \label{fig:disagreement_example}
\end{figure*}

Even in this restricted setting, different explanation methods (explainers) often provide contradictory explanations, such as in figure \ref{fig:disagreement_example}. The selection of incorrect or intentionally misleading explanations can misinform users and regulators, reinforcing systemic biases, and eroding public trust in AI systems \cite{disagreement_original, disagreement_new, rashomon}. Evaluating the quality of generated explanations is therefore a critically important problem. This is often done anecdotally -- of approximately 300 papers proposing new model explanation methods (explainers), one in three papers evaluated explanations anecdotally \cite{state_of_xai_evals}. Without consensus on the essential properties that explanations should possess and robust frameworks to numerically evaluate them, progress in the field remains slow, with fragmented benchmarks.

In this paper we do not propose a new XAI method but instead develop three general principles: \emph{local contextualization}, \emph{model relativism}, and \emph{on-manifold evaluation} to guide the evaluation of feature-importance explanations. When seeking to disambiguate the behavior of competing models from a Rashomon set, explanations should seek to highlight model differences rather than obscure them. This is operationalized through the \emph{model relativism} principle, which promotes explanations being dependent on the specific model used, rather than aligning towards an ideal value common to the entire Rashomon set. Conversely, explanations for models in a Rashomon set should be the same whenever the internal decision making mechanisms are the same. The \emph{on-manifold evaluation} principle requires explanations to depend only on on-manifold model behavior, discounting off-manifold behavior and promoting similar explanations for genuinely similar models.

\begin{figure}[h!]
    \centering
    \includegraphics[width=0.98\linewidth]{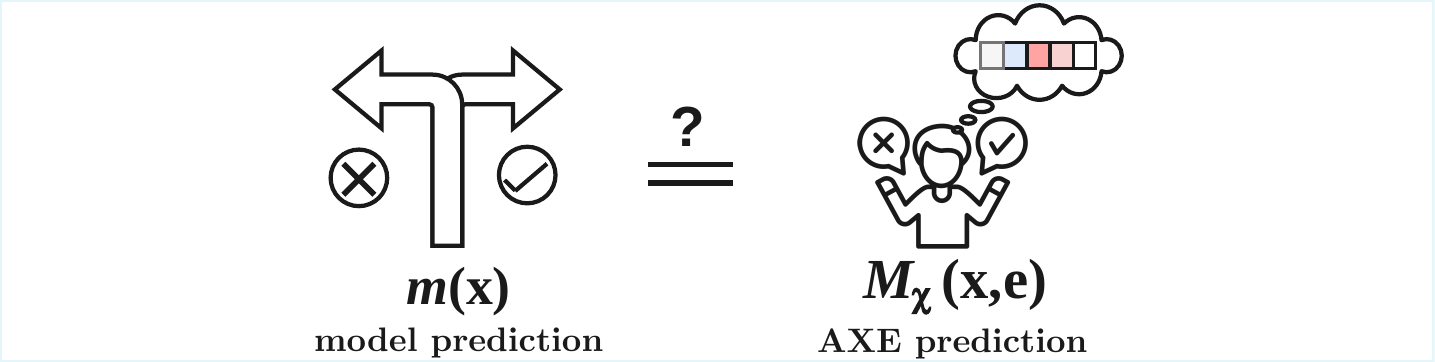}
    \caption{\textbf{Explanation evaluation}: \AXE evaluates the quality $q$ of explanation $\mathbf{e}$ by measuring how accurately prediction $m(\mathbf{x})$ can be recovered from dataset $\mathcal{X}$.}
    \label{fig:infographic2}
\end{figure}

We use these principles to propose \axe, a new feature sensitivity and ground-truth \textbf{Agnostic eXplanation Evaluation} framework that considers a good explanation to be one that correctly identifies the features most predictive of model outputs. \AXE is inspired by user research which indicates useful explanations are those that help users emulate and predict model behavior \cite{predictions_for_explanations}. \AXE satisfies the \emph{model relativism} principle, and can be used to distinguish models from a Rashomon set that make different predictions. It also simultaneously satisfies the \emph{on-manifold evaluation} principle, ascribing the same explanation to models in a Rashomon set if they do use the same internal decision mechanisms. Adversarial attacks have been proposed to exploit this latter vulnerability, where a discriminatory model's behavior is masked using another model that mimics discriminatory predictions (thus forming part of a Rashomon set), but varies in off-manifold behavior to mislead explainers into masking its discriminatory behavior. \AXE is invulnerable to this attack and is able to identify the correct explanation despite the adversarial manipulation.

\paragraph{Our notation: } For ease of exposition we specify our notation from figures \ref{fig:infographic1} and \ref{fig:infographic2}: input vector $\mathbf{x}$, model $m$, and explanation $\mathbf{e}$. We use these to define an explanation ``quality'' metric $q$ and an evaluation framework $Q$ here, and in algorithm \ref{alg:axe} we implement such a framework using \axe.

\begin{table*}[!b]
\centering
\caption{\textbf{Explanation Evaluation Metrics}: Definitions for ground-truth based explanation evaluation metrics (for explanation $\mathbf{e}$ and ground truth $\mathbf{e}^{*}$): \textbf{FA}, \textbf{RA}, \textbf{SA}, \textbf{SRA}, \textbf{RC} and \textbf{PRA} \cite{disagreement, agarwal2022openxai}; sensitivity based metrics \textbf{PGI} and \textbf{PGU} \cite{agarwal2022openxai, pgi_intro, pgi_counterfactuals}; and \axe. We list whether the metric satisfies each principle: Local Contextualization (\textbf{P1}), Model Relativism (\textbf{P2}), and On-Manifold Evaluation (\textbf{P3}).}

\begin{tabular}{lm{9.2cm}ccc}
\toprule
\textbf{Metric} & \textbf{Definition} & \textbf{P1} & \textbf{P2} & \textbf{P3} \\
\midrule

\makecell[l]{\textbf{FA:} Feature Agreement} & Fraction of top-n features common between $\mathbf{e}$ and $\mathbf{e}^{*}$. & \incorrect & \incorrect & \correct \\
\hdashline[1pt/2pt]
\makecell[l]{\textbf{RA:} Rank Agreement} & Fraction of top-n features common between $\mathbf{e}$ and $\mathbf{e}^{*}$ with the same position in respective rank orders. & \incorrect & \incorrect & \correct \\
\hdashline[1pt/2pt]
\makecell[l]{\textbf{SA:} Sign Agreement} & Fraction of top-n features common between $\mathbf{e}$ and $\mathbf{e}^{*}$ with the same sign. & \incorrect & \incorrect & \correct \\
\hdashline[1pt/2pt]
\makecell[l]{\textbf{SRA:} Signed Rank Agreement} & Fraction of top-n features common between $\mathbf{e}$ and $\mathbf{e}^{*}$ with the same sign and rank. & \incorrect & \incorrect & \correct \\
\hdashline[1pt/2pt]
\makecell[l]{\textbf{RC:} Rank Correlation} & Spearman's rank correlation coefficient for feature rankings from $\mathbf{e}$ and $\mathbf{e}^{*}$. & \incorrect & \incorrect & \correct \\
\hdashline[1pt/2pt]
\makecell[l]{\textbf{PRA:} Pairwise Rank Agreement} & Fraction of feature pairs for which relative ordering in $\mathbf{e}$ and $\mathbf{e}^{*}$ is the same. & \incorrect & \incorrect & \correct \\
\hdashline[1pt/2pt]
\makecell[l]{\textbf{PGI:} Prediction-Gap on Important\\Feature Perturbation} & Mean absolute change in model output upon perturbing top-n most important inputs. & \correct & \correct & \incorrect \\
\hdashline[1pt/2pt]
\makecell[l]{\textbf{PGU*:} Prediction-Gap on\\Unimportant Feature Perturbation} & Mean absolute change in model output upon perturbing top-n most unimportant inputs. & \correct & \correct & \incorrect \\
\hdashline[1pt/2pt] 
\makecell[l]{\textbf{\axe:} (ground-truth) Agnostic\\ eXplanation Evaluation} & Predictiveness of the top-n most important inputs in recovering model output. \textbf{Defined in algorithm \ref{alg:axe}.} & \correct & \correct & \correct \\

\bottomrule
\end{tabular}
\label{tab:prior_metrics}
\end{table*}

\begin{itemize}
    \item[-] \emph{Input Vector}: The input feature vector for any arbitrary datapoint is defined as:
\(
\mathbf{x} = [x_1, x_2, \dots, x_N] \in \mathbb{R}^N
\),
where \( N \) is the number of features, and each \( x_i \) is a real-valued feature.
    \item[-] \emph{Model and Prediction}: The model \( m \) is a mapping from the feature space to a binary output:
\(
m: \mathbb{R}^N \to \{0, 1\}
\),
and the prediction for input \( \mathbf{x} \) is given by:
\(
m(\mathbf{x}) = y_\text{pred} \in \{0, 1\}.
\)
    \item[-] \emph{Explanation}: A local feature importance explanation is denoted \( \mathbf{e} \). It is a function of the input \( \mathbf{x} \) and model \( m \) (implicitly model prediction \( m(\mathbf{x}) \) too). For an explainer $\mathcal{E}$,
\(
\mathbf{e} = \mathcal{E}(\mathbf{x}, m)
\), 
where \( \mathbf{e} \in \mathbb{R}^N \), and each component \( e_i \) represents the (signed) contribution or importance of the feature \( x_i \) to the prediction \( m(\mathbf{x}) \).
    \item[-] \emph{Explanation Quality Metric}: For dataset $\mathcal{X}$, the explanation quality metric \( q \in [0,1] \) evaluates the quality of explanation \( \mathbf{e} \) for a specific input \( \mathbf{x} \) and model \( m \). As a function,  $q=q_\mathcal{X}(\mathbf{x},m,\mathbf{e})$ where $0 \leq q \leq 1$ (greater $q$ is better). Previous work often refers to quality scores as fidelity or explanation faithfulness \cite{electronics8080832, abacus_xai_bench, gilpin_xai_review}.
\end{itemize}

An \emph{explanation evaluation framework} is a tuple \( (\mathcal{X}, m, \mathcal{E}, Q) \), where:
\begin{itemize}
    \item \( \mathcal{X} \in \mathbb{R}^{\nu \times N}\) is the dataset of inputs with $N$ features and $\nu$ datapoints, \( \mathcal{X} = \{\mathbf{x}_1, \mathbf{x}_2, \dots, \mathbf{x}_\nu \} \).
    \item \( m : \mathbb{R}^N \to \{0, 1\}\) is the model being explained.
    \item \( \mathcal{E} : \mathbb{R}^N \to \mathbb{R}^N \) is the explanation method that generates explanation \( \mathbf{e} \in \mathbb{R}^N \) for each datapoint \( \mathbf{x} \in \mathbb{R}^N \).
    \item \( Q \) is the aggregate quality score over the dataset computed as an average of explanation quality $q$:
\end{itemize}
    \[
    Q(\mathcal{X}, m, \mathcal{E}) = \frac{1}{\nu} \sum_{i=1}^\nu q(\mathbf{x}_i, m, \mathcal{E}(\mathbf{x}_i, m),\mathcal X).
    \]

\section{Three Principles for Evaluating Explanations}

Variations in explanations occur for many reasons. For example: (a) different input datapoints $\mathbf{x_1} \neq \mathbf{x_2}$ typically have different explanations; (b) different prediction models $m_1 \neq m_2$ typically have different explanations; and (c) as seen in figure  \ref{fig:disagreement_example}, explanations from different explainers can have different explanations due to varying off-manifold input sensitivity of the model $m$ in the neighborhood of $\mathbf{x}$. An evaluation framework that cannot distinguish between explanations from these varying scenarios and always scores diverse explanations the same is not helpful. We characterize these situations respectively with the following principles:

\begin{enumerate}
    \item \textbf{Local Contextualization}: \emph{Explanations should depend on the datapoint being explained.} For local explanations, when the datapoint $\mathbf{x}$ changes, the evaluation metric $q$ should not always prefer that the corresponding explanation $\mathbf{e}$ remain unchanged. Model behavior is not always identical across the data distribution.
    \item \textbf{Model Relativism}: \emph{Explanations should depend on the model being explained.} When selecting a different model $m$ from the Rashomon set that makes similar predictions but uses a different internal mechanism, the evaluation metric $q$ should promote corresponding changes in the explanation $\mathbf{e}$. Explanations should not hide, but surface behavioral differences across models in a Rashomon set.
    \item \textbf{On-manifold Evaluation}: \emph{Explanations on-manifold should not depend on changes in off-manifold model behavior.} When off-manifold model predictions $m(\mathbf{x}+\delta\mathbf{x})$ change, the evaluation metric $q$ should remain unchanged for explanation $\mathbf{e}$. Different models from a Rashomon set have the same behavior on the data manifold, but might vary in the predictions made ``off-manifold''. Explanations should be similar for such models, and evaluation metrics should consequently be independent of model behavior in ``explanation manifolds''.
\end{enumerate}

\begin{figure*}[!b]
    \centering
    \subfloat[Rank Agreement: $RA_{n=2}$]{
        \includegraphics[width=0.32\linewidth]{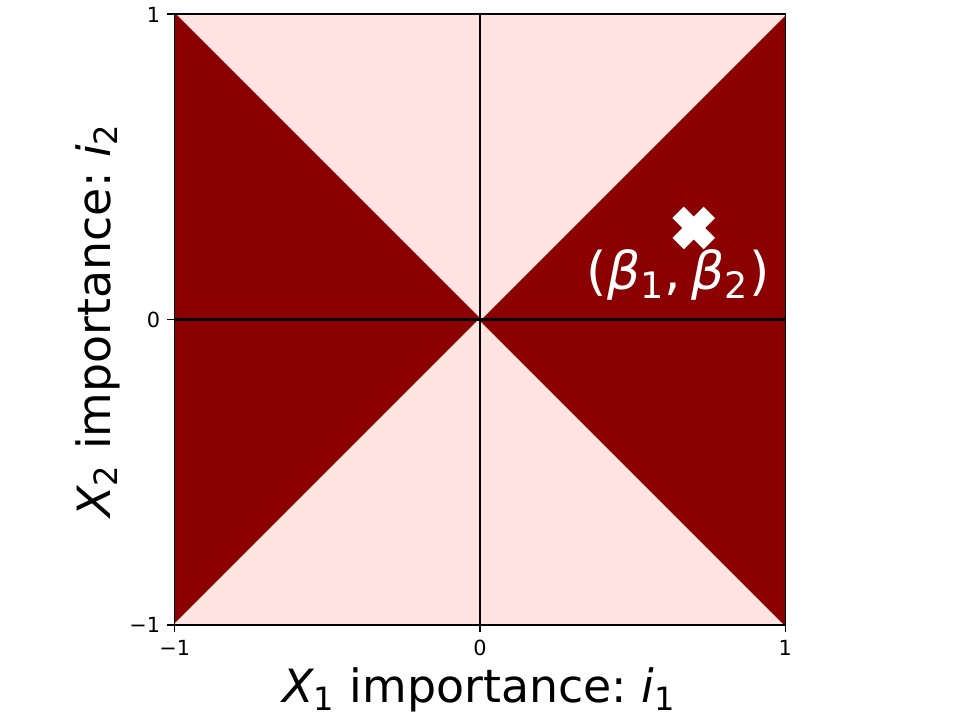}
    }
    \hfill
    \subfloat[Sign Agreement: $SA_{n=2}$]{
        \includegraphics[width=0.32\linewidth]{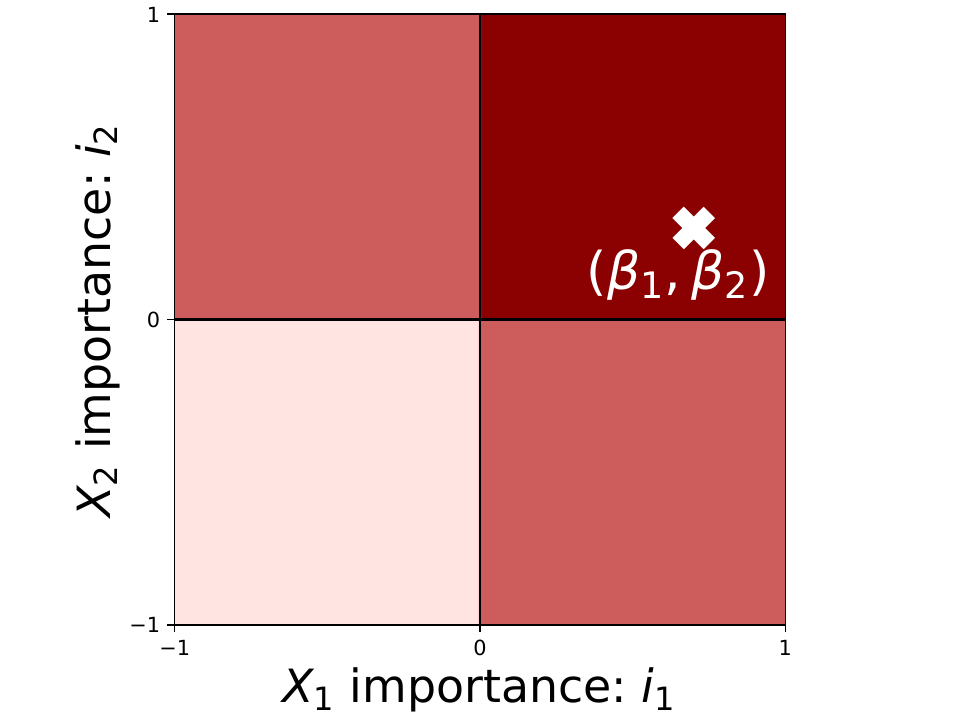}
    }
    \hfill
    \subfloat[Signed Rank Agreement: $SRA_{n=2}$]{
        \includegraphics[width=0.32\linewidth]{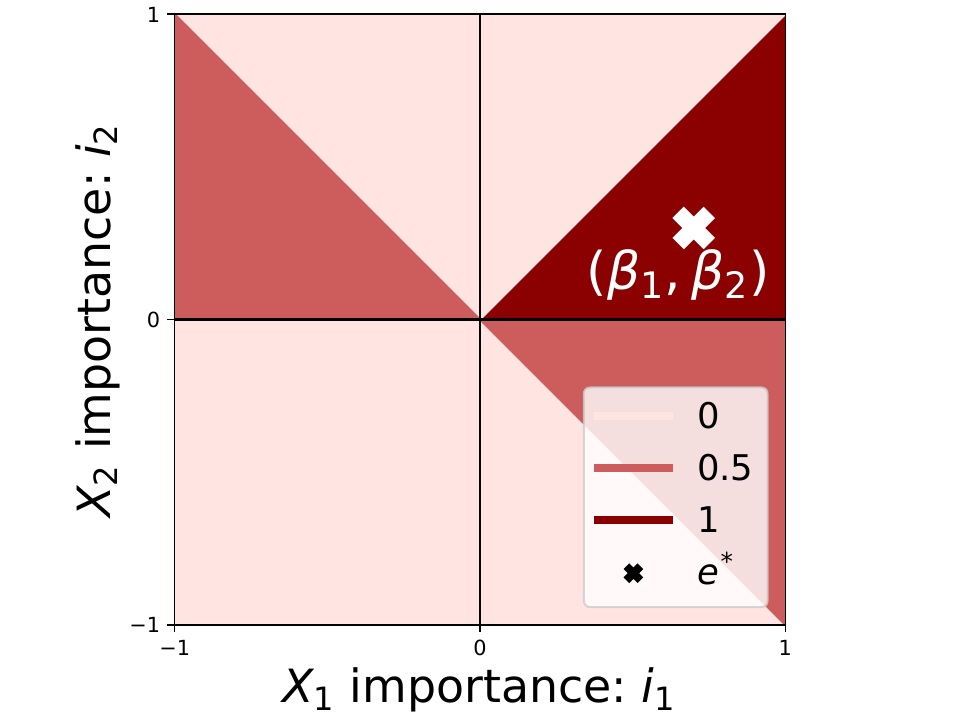}
    }
    \caption{\textbf{Explanation Quality is Invariant to Model Selection within the Rashomon Set}: Plots showing explanation quality $q$ (color) across $i_1$ and $i_2$ values for explanation $\mathbf{e} = (i_1, i_2)$. Model $m(\mathbf{x})=\beta_0 + \beta_1 X_1 + \beta_2 X_2$ has ground-truth $\mathbf{e}^{*} = (\beta_1, \beta_2) = (0.7,0.3)$. Consider model $m'(\mathbf{x})=\beta'_0 + \beta'_1 X_1 + \beta'_2 X_2$ that makes ``mostly'' similar predictions and belongs to the same Rashomon set $\{m, m'\} \subseteq \mathcal{M}$. If $\beta'_1 > \beta'_2 > 0$, then these explanation quality measurements remain unchanged (despite changes in ground-truth $\mathbf{e}^{*}$ against which explanation $\mathbf{e}$ will be compared). This violates the \emph{model relativism} principle.}
    \label{fig:gt_failures}
\end{figure*}

The \emph{on-manifold evaluation} principle is motivated by the observation that many explanation methods are variants of sensitivity analysis that capture how much synthetically varying a particular feature alters model outputs \cite{sens_in_xai, ood_sens_in_xai, efficient_sens_in_xai}. Ideally, an explanation for model behavior on datapoint $\mathbf{x_1}$ should not depend on model behavior on a different datapoint $\mathbf{x_2}=\mathbf{x}+\delta\mathbf{x}$. Further, evaluation frameworks that capture the fidelity of explanations with respect to synthetic neighborhoods around real points, are simply encoding a particular choice of sensitivity analysis without meaningfully evaluating the explanation quality. 

Broadly, prior explanation evaluation metrics can be classified into two types: 

\begin{enumerate}
    \item \textbf{Ground-truth} based  metrics compare the generated explanations $\mathbf{e}$ with ground-truth annotations $\mathbf{e}^{*}$, either collected by humans or inferred using a different proxy \cite{rise, pert_xai_1, pert_xai_2, gradcam}. These include Feature Agreement (FA), Sign Agreement (SA), Rank Agreement (RA), Signed Rank Agreement (SRA), Rank Correlation (RC) and Pairwise Rank Agreement (PRA) \cite{disagreement}.
    \item \textbf{Sensitivity} based metrics verify model sensitivity to the inputs declared important by an explanation \cite{rise, efficient_sens_in_xai, xai_visual_eval, xrai}. These have been summarized as Prediction Gap on Important Feature Perturbation (PGI) and Prediction Gap on Unimportant Feature Perturbation (PGU) \cite{pgi_intro, pgi_counterfactuals, agarwal2022openxai}.
\end{enumerate}

Sensitivity based metrics overindex on particular explanation methods. For instance, by defining the perturbation method appropriately, PGI can be made to emulate SHAP or LIME exactly, thus scoring particular explainers perfectly. For ground-truth based metrics, there can be similar variations in the notion of ``ground-truth''. Proposals include one ``ground-truth'' per datapoint $\mathbf{x}$, unintentionally introducing independence from $m$ \cite{pointinggame, salsanity} violating the \emph{model relativism} principle. Consider a datapoint $x$ such that all models in a Rashomon set $m_i \in \mathcal{M}$ produce an identical prediction. By having an immutable ``ground-truth'' $\mathbf{e}^{*}$ to compare diverse explanations $e$ with, the evaluation metric incentivizes explainers to output the same explanation regardless of the particular model $m_i$. 

With images especially, an explanation is often considered good if it selects the ``correct'' region as important in an image -- regardless of whether the model used those features \cite{pointinggame, salsanity, psycho_pointing_evaluation}. This leads to egregious violations of the model relativism principle, as many models $m_i$ detecting the same object in an image (but due to different pixels) can form a Rashomon set $\mathcal{M}$ \cite{salsanity, pointinggame, psycho_pointing_evaluation}. \AXE is inspired by methods proposed to evaluate such explanations \cite{axe_in_images, psycho_pointing_evaluation}. 

\section{Ground-truth Metrics promote the same Explanation Across Models in a Rashomon Set}
\label{subsec:gt_failures}

Real world situations lack access to an oracle to provide ground-truth explanations $\mathbf{e}^{*}$ \cite{electronics10050593, missing_ground_truths}. For linear models, a common resolution adopts the model coefficients as the ``ground-truth'' for all datapoints $\mathbf{x} \in \mathcal{X}$ in a dataset \cite{disagreement, agarwal2022openxai}. Figure \ref{fig:gt_failures} depicts such an example, showcasing a violation of the \emph{model relativism} principle.

Consider a model $m$ with two input features, $X_1$ and $X_2$, which makes predictions $y$, parameterized $y = \beta_0 + \beta_1 X_1 + \beta_2 X_2$. A related model $m'$ is part of the same Rashomon set $\{m, m'\} \subseteq \mathcal{M}$, has slightly different decision boundaries but makes similar predictions $y \approx y' = \beta'_0 + \beta'_1 X_1 + \beta'_2 X_2$. For datapoint $\mathbf{x}$ and model $m$ feature importances are denoted $i_1$ and $i_2$, with explanation $e = [i_1, i_2]$. FA, RA, SA, SRA, RC, and PRA measure explanation quality by comparing with ground-truth explanation $\mathbf{e}^{*} = [\beta_1, \beta_2]$ \cite{disagreement}.
This comparison takes many forms, with definitions provided in table \ref{tab:prior_metrics}. For example, our $N=2$ feature setup implies that for the top $n$ features: $\text{FA}_{n=0} = 0$, $\text{FA}_{n=1} \in \{0, 0.5, 1.0\}$, and $\text{FA}_{n=2} = 1$, and that $\text{FA}_{n=1} = \text{RA}_{n=2} = \text{PRA}_{n=2}$, while RC is undefined.

Plotting evaluation metric $q$ for all possible explanations $e = [i_1, i_2]$, model $m$ with $\beta_1 = 0.7$ and $\beta_2 = 0.3$ (so $\beta_1 > \beta_2 > 0$), we see that regardless of the specific value of $\mathbf{e}$, there are regions where the resulting $RA_{n=2}$ is the same (figure \ref{fig:gt_failures} a). Similarly, the $SA_{n=2}$ is the same across $i_1, i_2$ regions (figure \ref{fig:gt_failures} b) and $SRA_{n=2}$ too (figure \ref{fig:gt_failures} c). Concretely, any explanation $e = (i_1, i_2)$ such that $i_1 > 0$, $i_2 > 0$, and $i_1 > i_2$ (region labeled 1 in figure \ref{fig:gt_failures} c) is guaranteed to have the same FA, RA, SA, SRA, RC, and PRA.

Any model $m' \in \mathcal{M}$ with similar predictions such that $m \neq m'$ with the same explanation quality evaluations as model $m$ from figure \ref{fig:gt_failures} would violate the \emph{model relativism} principle.
Upon observation, all models $m'$ such that $\beta'_1 > \beta'_2 > 0$ satisfy this condition. The relative importance of $X_1$ with respect to $X_2$ can range from $1$ to $\infty$ in the limit, but the ground-truth based explanation quality metrics FA, RA, SA, SRA, RC, and PRA would be unchanged. For instance if $m'$ has similar predictions on the test set to $m$ with $\beta'_1 = 0.5$ and $\beta'_2 = 0.3$, then the models belong to the same Rashomon set but the evaluation promotes the exact same explanation for both models, despite the relative importance of the features being different.
This example demonstrates \emph{local contextualization} violations too. The presence of entire regions (as opposed to individual points) in figure \ref{fig:gt_failures} with a common explanation quality value $q$ indicates that different possible explanations $\mathbf{e}$ map to the same quality of $1$, $0.5$, or $0$. Since the comparison is always with a fixed vector $\mathbf{e}^{*}$ which does not change with datapoint $\mathbf{x}$, explanations for different datapoints are incentivized to be identical, promoting a single ``global'' explanation over per-datapoint ``local'' explanations.

\section{A Sensitivity and Ground-Truth Agnostic Explanation Evaluation Framework}

Inspired by desiderata from user studies \cite{predictions_for_explanations, user_predictiveness_1, user_predictiveness_2, user_predictiveness_3, user_predictiveness_4, user_predictiveness_5}, and seeking to alleviate common problems with ground-truth based evaluation, we consider a simple alternative to sensitivity-driven methods of XAI evaluation: the important features in any explanation should be more predictive of the model output than the unimportant features. \AXE adopts classifier accuracy \cite{axe_in_images} to measure the predictiveness of the top-n important features. This ``top-n'' style formulation, just like prior metrics from table \ref{tab:prior_metrics}, is considered intuitive for practitioners \cite{agarwal2022openxai}. 

For datapoint $\mathbf{x}$ and explanation $\mathbf{e}$, the top-n most important features are the importances with the largest absolute values. To measure explanation quality $q$, \AXE uses predictiveness -- the accuracy of a k-Nearest Neighbors (\textit{k}-NN) model $M^k$ in recovering the model prediction $m(\mathbf{x})$ using only the subset of the the top-n most important features. The \textit{k}-NN $M$ mimics the prediction $m(\mathbf{x})$ of model $m$ by averaging over the predictions from the \textit{k} neighbors nearest to $\mathbf{x}$ \cite{knn1, knn2}.

\begin{algorithm}[ht]
\caption{Evaluating Explanation Quality with $\bm{\text{\axe}_n^k}$}
\label{alg:axe}
\begin{algorithmic}[1]
\REQUIRE Number of Features $n$, Number of Neighbors $k$ \\ Dataset $\mathcal{X} = \{\mathbf{x}_i\}_{i=1}^{\nu}$, Predictions $Y_\text{preds} = \{y_i\}_{i=1}^{\nu}$, and Explanations $E = \{\mathbf{e}_i\}_{i=1}^{\nu}$
\STATE Initialize an empty list: $\hat{Y} \gets []$
\FOR{each datapoint $\mathbf{x}_i$ and explanation $\mathbf{e}_i$ in $(\mathcal{X}, E)$}
    \STATE Find the $n$ most important input features: $f_\text{imp} \gets \text{ImpFeatures}(\mathbf{e}_i, n)$
    \STATE Create $\mathcal{X}_f$ with subset of features $f_{imp}$ from $\mathcal{X}$
    \STATE Train K-NN model $M^k_i$ with inputs $\mathcal{X}_f$ and target $Y_\text{preds}$
    \STATE Obtain prediction $\hat{y}_i$ from $M^k_i$ for datapoint $\mathbf{x}_i$
    \STATE Append $\hat{y}_i$ to $\hat{Y}$
\ENDFOR
\STATE Return performance measure: $ \text{Accuracy}(\hat{Y}, Y_\text{preds})$
\end{algorithmic}
\end{algorithm}

Algorithm \ref{alg:axe} defines our framework for evaluating explanations. We denote the target variable \(Y\), predicted from the input dataset \(\mathcal{X}\) consisting of $\nu$ datapoints and $N$ features. The model \(m\) makes predictions \(Y_\text{preds} = m(\mathcal{X})\), for which a set of feature-importance explanations \(E\) can be computed such that $\exists \mathbf{e}_i \in E  \forall  \mathbf{x}_i \in \mathcal{X}$. \AXE fits multiple \textit{k}-NN models $M^k_i$ to predict model outputs $Y_\text{preds}$, not data labels $Y$. \AXE has two hyperparameters: \(n\) for the ``top-n'' number of important features to use and \(k\), for the number of neighbors to use in the \textit{k}-NN model $M^k$; denoted \(\text{\axe}_n^k \).

It is essential \emph{not} to use the same \textit{k}-NN model for all predictions -- \AXE does not build a global \textit{k}-NN surrogate to measure explanation quality. For each datapoint $\mathbf{x}_i$, we use a unique \textit{k}-NN model that considers the top-n most important features for that particular explanation $\mathbf{e}_i$. This insight is critical to ensure that \AXE does not just report the accuracy of an arbitrary \textit{k}-NN model over the entire dataset.
Consequently, \AXE satisfies the \emph{local contextualization} principle by using a different set of neighbors for each datapoint $\mathbf{x}$. Each datapoint has its own nearest-neighbors model \(M_i^k\), unique to each prediction \(\hat{y}_i\) for $\mathbf{x}_i$. It further satisfies the \emph{model relativism} principle by training model \(M_i^k\) to predict the classifier response \(Y_\text{preds}\), rather than the target feature \(Y\), making the quality metric $q$ dependent on the model $m$. Finally, \AXE satisfies the \emph{on-manifold evaluation} principle because the \textit{k}-NN models are explicitly limited to the existing data manifold and do not rely on new datapoints $\mathbf{x} \notin \mathcal{X}$, avoiding feature sensitivity measures.

\section{Detecting Explanations Fairwashed Using Different Models from a Rashomon Set}

\begin{table*}[ht!]
\caption{
\textbf{Detecting explanation fairwashing}: We replicate an adversarial fairwashing attack \cite{adv_attack} and generate spurious explanations, which we try to then detect using quality metrics $q$ that do not need ground-truths: \axe, PGI, and PGU. $E_\rho$ is a set of explanations that correctly denote that the most important model input feature is $X_\rho$. $E_\phi$ and $E_\psi$ are explanations where the most important feature is $X_\phi$ or $X_\psi$. We are explaining model $m_e$ which uses the same internal mechanism as model $m$ for all datapoint in the data input manifold, ie $\{m, m_e\} \subseteq \mathcal{M}$ and both use feature $X_\rho$ to make predictions.
A good evaluation metric $q$ should distinguish manipulated explanations from correct explanations: (i) \(\bar{q}(E_\rho) > \bar{q}(E_\phi) \) and (ii) \(\bar{q}(E_\rho) > \bar{q}(E_\psi) \); where $\bar{q}(E) = \sum_{\mathbf{e} \in E} q(\mathbf{e}) / |E|$. \textbf{\AXE always scores the correct explanation better than other explanations.}
}
\label{tab:pgiufails}
\centering
\begin{tabular}{ll| @{\hspace{0.1em}} r @{\hspace{2.5em}} r @{\hspace{3.5em}} r @{\hspace{2.5em}} r r @{\hspace{0.1em}} |c}
& & & & & & & \\
\toprule
\multirow{2}{*}{\thead{ \\ \textbf{Dataset} }} & 
\multirow{4}{*}{\thead{ \textbf{Adversarial} \\ \textbf{Model} }} & 
\multirow{2}{*}{ \thead{ \textbf{Eval.} \\ \textbf{Metric $q$} \\ (\(n=1\)) }} & 
\multicolumn{4}{l|}{\thead{\textit{Evaluating explanations with a single important attribute:}}} & 
\multirow{2}{*}{\thead{  \bm{$\bar{q}(E_\rho)>\bar{q}(E_\phi)$} \\ \textit{and} \\ \bm{$\bar{q}(E_\rho)>\bar{q}(E_\psi)$} }} \\

 & & & \thead{\textbf{Protected} \\ \bm{$\bar{q}(E_\rho)$} } & 
 \thead{\textbf{Foil 1} \\ \bm{$\bar{q}(E_\phi)$} } & 
 \thead{\textbf{Foil 2} \\ \bm{$\bar{q}(E_\psi)$} } & 
 \thead{\textbf{Other} \\ \bm{$\bar{q}(E_\omega)$} } \\

\specialrule{\lightrulewidth}{0.2em}{0pt}

\multirow{6}{*}{\makecell[l]{ \textbf{German} \\ \textbf{Credit} } }
 & \multirow{3}{*}{ \makecell[l]{ \(\bm{m_{L1}}\) \\ (LIME, 1 foil) } } 

 & PGI & 0.032 & 0.148 & na & 0.018 & \incorrect \\
 & & (-)PGU & -0.486 & -0.536 & na & -0.483 & \correct \\ 
 & & \axe & 1.000 & 0.680 & na & 0.617 & \correct \\ 

\cdashline{2-8}[1pt/2pt]
 & \multirow{3}{*}{ \makecell[l]{ \(\bm{m_{S1}}\) \\ (SHAP, 1 foil) } } 

 & PGI & 0.037 & 0 & na & 0.037 & \correct \\
 & & (-)PGU & -0.475 & -0.529 & na & -0.478 & \correct \\
 & & \axe & 0.990 & 0.690 & na & 0.622 & \correct \\

\hdashline[5pt/2pt]

\multirow{12}{*}{ \makecell[l]{ \textbf{COMPAS} } } 
 & \multirow{3}{*}{ \makecell[l]{ \(\bm{m_{L1}}\) \\ (LIME, 1 foil) } } 

 & PGI & 0.006 & 0 & na & 0.067 & \correct \\
 & & (-)PGU & -0.481 & -0.479 & na & -0.431 & \incorrect \\ 
 & & \axe & 0.992 & 0.739 & na & 0.534 & \correct \\ 

\cdashline{2-8}[1pt/2pt]
 & \multirow{3}{*}{ \makecell[l]{ \(\bm{m_{S1}}\) \\ (SHAP, 1 foil) } } 

 & PGI & 0.006 & 0.035 & na & 0.009 & \incorrect \\
 & & (-)PGU & -0.091 & -0.077 & na & -0.090 & \incorrect \\ 
 & & \axe & 0.968 & 0.761 & na & 0.527 & \correct \\ 

\cdashline{2-8}[3pt/2pt]
 & \multirow{3}{*}{ \makecell[l]{ \(\bm{m_{L2}}\) \\ (LIME, 2 foils) } } 

 & PGI & 0.006 & 0 & 0.001 & 0.075 & \correct \\
 & & (-)PGU & -0.520 & -0.520 & 0-0.524 & -0.464 & \incorrect \\
 & & \axe & 0.990 & 0.739 & 0.735 & 0.533 & \correct \\ 

\cdashline{2-8}[1pt/2pt]
 & \multirow{3}{*}{ \makecell[l]{ \(\bm{m_{S2}}\) \\ (SHAP, 2 foils) } } 

 & PGI & 0.005 & 0.039 & 0.041 & 0.010 & \incorrect \\
 & & (-)PGU & -0.104 & -0.090 & -0.092 & -0.106 & \incorrect \\
 & & \axe & 0.956 & 0.746 & 0.731 & 0.531 & \correct \\ 

 \hdashline[5pt/2pt]

\multirow{12}{*}{ \makecell[l]{ \textbf{Communities} \\ \textbf{and Crime} } } 
 & \multirow{3}{*}{ \makecell[l]{ \(\bm{m_{L1}}\) \\ (LIME, 1 foil) } } 

 & PGI & 0.103 & 0 & na & 0.029 & \correct \\
 & & (-)PGU & -0.479 & -0.460 & na & -0.481 & \incorrect \\ 
 & & \axe & 1.000 & 0.765 & na & 0.793 & \correct \\ 

\cdashline{2-8}[1pt/2pt]
 & \multirow{3}{*}{ \makecell[l]{ \(\bm{m_{S1}}\) \\ (SHAP, 1 foil) } } 

 & PGI & 0.089 & 0.006 & na & 0.005 & \correct \\
 & & (-)PGU & -0.446 & -0.429 & na & -0.448 & \incorrect \\ 
 & & \axe & 0.985 & 0.765 & na & 0.790 & \correct \\

\cdashline{2-8}[5pt/2pt]
 & \multirow{3}{*}{ \makecell[l]{ \(\bm{m_{L2}}\) \\ (LIME, 2 foils) } } 

 & PGI & 0.101 & 0.001 & 0.001 & 0.034 & \correct \\
 & & (-)PGU & -0.534 & -0.536 & -0.536 & -0.535 & \correct \\
 & & \axe & 0.995 & 0.760 & 0.760 & 0.792 & \correct \\

 \cdashline{2-8}[1pt/2pt]
 & \multirow{3}{*}{ \makecell[l]{ \(\bm{m_{S2}}\) \\ (SHAP, 2 foils) } } 

 & PGI & 0.094 & 0.006 & 0.005 & 0.008 & \correct \\
 & & (-)PGU & -0.479 & -0.470 & -0.479 & -0.479 & \incorrect \\
 & & \axe & 0.955 & 0.760 & 0.755 & 0.781 & \correct \\ 

\specialrule{\heavyrulewidth}{0pt}{0pt}

\end{tabular}
\end{table*}

We simulate a state-of-the-art adversarial attack \cite{adv_attack} on explanations in a real-world setting where ground-truths remain unknown. The attack modifies a model $m$, known to be discriminatory, creating new models $m_S$ or $m_L$ that respectively fool SHAP and LIME into generating explanations $\mathbf{e}$ that show the discriminatory feature as unimportant. The discriminatory decision procedure employed internally by model $m$ remains unchanged in models $m_S$ and $m_L$, and the three models make similar predictions. The models $m$, $m_S$, and $m_L$ therefore form a Rashomon set of models that make similar predictions.

This attack fairwashes model explanations by hiding discriminatory model behavior. Imagine explanations for the diabetes prediction model figure \ref{fig:disagreement_example} showing that the model used benign inputs to make its prediction, when the model actually made predictions using protected attributes that are medically irrelevant. Like $m$, $m_S$ and $m_L$ too make decisions using only the ``protected'' feature ($X_\rho$), but they fool explainers SHAP and LIME respectively into generating explanations showing spurious "foil" features ($X_\phi, X_\psi$) as the most important \cite{adv_attack}. A good evaluation metric would identify the same explanation to be of high quality for all three models. This explanation would ideally be the ``correct'' one indicating that a protected feature ($X_\rho$) was important to the model prediction. Models $m_S$ and $m_L$ are manipulated such that other explanations (showing features $X_\phi or X_\psi$ as important) get generated, but a good explanation evaluation metric $q$ should identify these manipulated explanations by scoring them poorly.

Formally, given a data manifold $\mathcal{D}$ and model $m$ that uses a ``protected'' input feature $X_\rho$ (eg. race or gender) to make predictions, the attack enumerates models $m_e$ in the Rashomon set of the model $\{m, m_e\} \subseteq \mathcal{M}$ such that they make the same predictions using the same protected input feature on the data manifold, that is $m_e(\mathbf{x}) \approx m(\mathbf{x}) \forall \mathbf{x} \in \mathcal{D}$. The model $m_e$ might differ from $m$ off-manifold, and in the case of $m_S$ and $m_L$ is explicitly designed such that the behavior outside of data manifold $\mathcal{D}$, in the explanation manifold of SHAP or LIME, presents spurious ``foil'' input features $X_\phi$ or $X_\psi$ as most important.

Model behavior differs only for datapoints that lie off the data manifold $x \notin \mathcal{D}$. If we were considering these datapoints, where models genuinely used different features (eg. $X_\phi$ vs $X_\psi$), then \emph{model relativism} would require explanations to differ: explaining $m$ should indicate protected input feature $X_\rho$ as important, while explaining $m_e$ should ascribe importance to $X_\phi$ or $X_\psi$. However, when considering datapoints on-manifold $\mathbf{x} \in \mathcal{D}$, \emph{on-manifold evaluation} requires feature importance explanations for $m_e$ to correctly identify the protected attribute $X_\rho$ as important. Recall that for these datapoints, both $m$ and $m_e$ use the same internal mechanism, making identical predictions using the protected feature $X_\rho$.

We use the same datasets and models as the original adversarial attack \cite{adv_attack}: the German Credit dataset from lending \cite{german_credit} and the COMPAS \cite{compas} dataset and the Communities and Crime dataset \cite{communities_and_crime} from criminal justice. We then manually construct competing feature-importance explanations: $E_\rho$ is the set of all explanations where the protected feature $X_\rho$ is the most important feature. $E_{\phi}$ and $E_{\psi}$ are the explanation sets where the foil feature $X_\phi$ or $X_\psi$ is the most important, respectively. Finally, $E_\omega$ is the set of all other explanations, where the most important feature is neither protected nor a foil feature. The credit model is adversarially modified to deliberately favor men over others, and the criminal justice models are modified to favor white people over others. Explanations that reveal this ($E_\rho$) are correct. Explanations that mask this ($E_{\phi},E_{\psi}$) are spurious. Evaluation metrics $q$ should identify this by the condition $\bar{q}(E_{\rho}) > \bar{q}(E_{\phi})$ and $\bar{q}(E_{\rho}) > \bar{q}(E_{\psi})$; where $\bar{q}$ is the average quality $\bar{q}(E) = \sum_{\mathbf{e} \in E} q(\mathbf{e}) / |E|$.

We test \AXE against PGI and (-)PGU (PGU values flipped negative so that higher scores are better like PGI and \axe) by scoring various explanations for models $m_e$. We use the German Credit \cite{german_credit} dataset from lending and the COMPAS \cite{compas} and Communities and Crime \cite{communities_and_crime} datasets from criminal justice. We assume the adversary is interested in deploying binary classifier $m$ that makes predictions using protected characteristics $X_\rho$ race or gender. We consider 4 kinds models in the Rashomon set: designed to fool explainers SHAP and LIME, and with 1 or 2 spurious foil features respectively, giving us $\mathcal{M} = \{ m_{S1} , m_{S2} , m_{L1} , m_{L2} \}$. We then score competing explanations for each of these models, checking that explanations identifying the protected feature $X_\rho$ as most important receive the highest scores.


Table \ref{tab:pgiufails} we summarizes our results. To test explanations constructed to consider one feature as important and all others as unimportant, we use \(\text{\AXE}_{n=1}\),  \(\text{PGI}_{n=1}\) and  \(\text{PGU}_{n=1}\). The last column \ref{tab:pgiufails} shows PGU failing to discern correct explanations $E_\rho$ from spurious ones $E_\psi$, $E_\phi$ 7 out of 10 times, and PGI failing 3 out of 10 times. The overall error rate for PGI and PGU is 50\%, while \AXE has no failures yielding an error-rate of 0\%. Off-manifold manipulation by models $m_e$ from the Rashomon set of the model $m \in \mathcal{M}$, designed to mislead explainers SHAP and LIME, also misleads sensitivity based evaluation metrics PGI and PGU.

\section{Conclusion}
\label{sec:conclusion}

In this paper we have focused on two important aspects of XAI as they relate to the Rashomon effect. If models make similar predictions but through different internal mechanisms, explanations should seek to highlight these differences (\emph{model relativism}). Figure \ref{fig:gt_failures} presents an example case where prior ground-truth based metrics fail this simple test, and promote a single explanation across models in a Rashomon set. Conversely, if models in a Rashomon set use the same internal mechanism to make a prediction, differing only off-manifold, then explanations on-manifold should be the same for all models (\emph{on-manifold evaluation}). For models from a Rashomon set that have the exact same internal mechanisms, if explanations generated are different due to adversarial fairwashing then this should be detected when evaluating them. Our experiment in table \ref{tab:pgiufails} shows that unlike prior metrics, our proposed metric \AXE can detect adversarial fairwashing because it follows the three principles of explanation evaluation. While this abridged paper highlights connections between explanation evaluation and predictive multiplicity caused by the Rashomon effect, additional experiments and theory situating \AXE in the AI explainability literature can be found in the original paper ``Evaluating Model Explanations without Ground-truth''.

\section{Acknowledgments}

This research was supported by the Wellcome Trust (grant no. 223765/Z/21/Z), Sloan Foundation (grant no. G-2021-16779), Department of Health and Social Care, EPSRC (grant no. EP/Y019393/1), and Luminate Group. This work has also been supported by the Alexander von Humboldt Foundation in the framework of the Alexander von Humboldt Professorship (Humboldt Professor of Technology and Regulation) endowed by the Federal Ministry of Education and Research via the Hasso Plattner Institute. The donors had no role in the decision to publish or the preparation of this paper.


\bigskip

\bibliography{aaai2026}


\end{document}